%% file: main.tex
\documentclass{article}
\pdfoutput=1

\PassOptionsToPackage{numbers, compress}{natbib}

\usepackage[final]{neurips_2022}

\usepackage[utf8]{inputenc} 
\usepackage[T1]{fontenc}    
\usepackage{hyperref}       
\usepackage{url}            
\usepackage{booktabs}       
\usepackage{amsfonts}       
\usepackage{nicefrac}       
\usepackage{microtype}      
\usepackage[dvipsnames]{xcolor}
\usepackage[noend]{algorithmic}
\usepackage{algorithm}
\usepackage{tikz}
\usepackage{wrapfig}
\usepackage{caption}
\usepackage{subcaption}
\usepackage{graphicx}
\usepackage{amsmath}
\usepackage{enumitem}

\input{math_commands.tex}

\title{Generalization Gap in Amortized Inference}

\author{%
   Mingtian Zhang \And Peter Hayes \And David Barber 
   \Aff
 Centre for Artificial Intelligence, University College London\\
  \texttt{\{m.zhang,p.hayes,d.barber\}@cs.ucl.ac.uk} \\
}

\begin{document}

\maketitle

\begin{abstract}
The ability of likelihood-based probabilistic models to generalize to unseen data is central to many machine learning applications such as lossless compression. In this work,  we study the generalization of a popular class of probabilistic model - the Variational Auto-Encoder (VAE). We discuss the two generalization gaps that affect VAEs and show that overfitting is usually dominated by amortized inference. Based on this observation, we propose a new training objective that improves the generalization of amortized inference. We demonstrate how our method can improve performance in the context of image modeling and lossless compression.
\end{abstract}

\section{Introduction}

Probabilistic models have achieved great success in many machine learning applications~\cite{bishop2006pattern,barber2011bayesian}. Given a set of training data that are sampled from an underlying data distribution  $\mathcal{X}_{train}=\{x_1,\cdots, x_N\} \sim p_d(x)$, the goal of probabilistic modelling is to approximate $p_d(x)$ with a model $p_\theta(x)$. A principled method to learn   $\theta$ is to minimize the  Kullback-Leibler (KL) divergence 
\begin{align}
    \mathrm{KL}(p_d(x)||p_\theta(x))=\langle\log p_d(x)\rangle_{p_d(x)}-\langle\log p_\theta(x)\rangle_{p_d(x)},
\end{align}
where we use $\langle\cdot\rangle$ to denote integration: $\langle f(x)\rangle_{p(x)}\equiv\int f(x)p(x)dx$. The first term  represents the negative entropy  of the data distribution $-H(p_d)\equiv \langle\log p_d(x)\rangle_{p_d(x)}$, which is a constant. The second cross entropy term involves the integration over the unkown data distribution $p_d(x)$, which can be approximated 
by the Monte-Carlo approximation using the training dataset $\mathcal{X}_{train}$
\begin{align}
    \langle\log p_\theta(x)\rangle_{p_d(x)}\approx\frac{1}{N}\sum_{n=1}^N \log p_\theta(x_n).
\end{align}
Therefore, estimating $\theta$ by minimizing the KL divergence is equivalent to Maximum Likelihood Estimation (MLE) when $N\rightarrow \infty$.

For a finite dataset, a common concern in both supervised and unsupervised learning is that the probabilistic model may overfit to the training dataset $\mathcal{X}_{train}$, degrading generalization performance~\cite{shalev2014understanding}. The generalization performance in the unsupervised setting can be measured by the test likelihood~\cite{zhang2021ood}: $\frac{1}{M}\sum_{n=1}^{M} \log p_\theta (x'_m)$, where $\mathcal{X}_{test}=\{x'_1,\ldots, x'_M\} \sim p_d(x)$ is the test dataset. A model that has overfit to the training dataset $\mathcal{X}_{train}$ generally results in a high training likelihood but a low test likelihood.
Although the test likelihood is a common evaluation criterion~\cite{theis2015note}, the factors that affect the generalization of unsupervised probabilistic models are less well studied in comparison to supervised learning.
We posit that this is because for common tasks, like sample generation or representation learning, good generalization in terms of the test likelihood is not a sufficient measure of performance. For example implicit models can generate sharp samples without having a likelihood function \cite{goodfellow2014generative,arjovsky2017wasserstein,zhang2020spread} and representations learned by latent variable models can be arbitrarily transformed without changing the likelihood~\cite{locatello2019challenging}. However, in recent applications that use deep generative models for lossless compression~\cite{townsend2019practical,townsend2019hilloc,kingma2019bit,zhang2021ood,zhang2022parallel}, generalization in terms of the test likelihood directly indicates higher compression rate~\cite{zhang2021ood}. Specifically,
given a probabilistic model $p_\theta(x)$, a lossless compressor can be constructed to compress a test data point $x'$ to a bit string with length approximately equal to $-\log_2 p_\theta(x')$. When $p_\theta(x)\rightarrow p_d(x)$, the average compression length attains the entropy of the data distribution $-\frac{1}{M}\sum_{m=1}^M\log_2 p_\theta(x_m') \rightarrow H(p_d)$, which is \emph{optimal} under Shannon's source coding theorem~\cite{shannon2001mathematical}, see Appendix~\ref{app:bbans} for a detailed introduction. Therefore, a better test likelihood can lead to a greater saving in bits and so understanding and improving generalization of deep generative models is an important challenge.

\subsection{Variational Auto-Encoder}
A popular type of probabilistic model is the Variational Auto-Encoder (VAE)~\citep{kingma2013auto,rezende2014stochastic}, which assumes a latent variable model
$p_\theta (x)=\int p_\theta(x|z)p(z)dz$.
For a nonlinear parameterization of $p_\theta(x|z)$ (e.g. a deep neural network), the evaluation of $\log p_\theta(x)$ involves solving an intractable integration over $z$. In this case,
the evidence lower bound (ELBO) can  be used to side-step the intractability
\begin{align}
    \langle\log p_\theta(x)\rangle_{p_d(x)} &\geq \langle\log p_\theta(x,z)-\log q_\phi(z|x)\rangle_{q_\phi(z|x)p_d(x)}\equiv \big\langle\mathrm{ELBO}(x,\theta,\phi)\big\rangle_{p_d(x)},\label{eq:elbo}
\end{align}
where $q_\phi(z|x)$ is a variational posterior  parameterized by a neural network with parameter $\phi$. The use of an approximate posterior of the form $q_\phi(z|x)$ is called \emph{amortized inference}. 
To better understand this objective, we can rewrite the expected ELBO as the following
\begin{align}
     \big\langle\mathrm{ELBO}(x,\theta,\phi)\big\rangle_{p_d(x)}&= \Big\langle \log p_\theta(x)-\mathrm{KL}(q_\phi(z|x)||p_\theta(z|x))\Big\rangle_{p_d(x)}\label{eq:rewrite:ELBO:0}
\\&=-\underbrace{H(p_d)}_{const.}-\underbrace{\mathrm{KL}(p_d(x))||p_\theta(x))}_{\text{model learning}}-\underbrace{\big\langle\mathrm{KL}(q_\phi(z|x)||p_\theta(z|x))\big\rangle_{p_d(x)}}_{\text{amortized inference}},
    \label{eq:rewrite:ELBO}
\end{align}
We  denote the posterior family of $q_\phi(z|x)$ as $\mathcal{Q}$, which is indexed by a finite dimensional $\theta$~\cite{wang2019frequentist}. If $\mathcal{Q}$ is flexible enough such that the true posterior $p_\theta(z|x)\in \mathcal{Q}$, where $p_\theta(z|x)\propto p_\theta(x|z)p(z)$, then in the optimum of Equation \ref{eq:rewrite:ELBO:0},  we have $\mathrm{KL}(q_\phi(z|x)||p_\theta(z|x))=0\Rightarrow q_\phi(z|x)=p_\theta(z|x)$ for $x\sim p_d(x)$ and the ELBO will be equal to the log-likelihood $\mathrm{ELBO}(x,\theta,\phi)=\log p_\theta(x)$~\cite{kingma2013auto,blei2017variational}.  Many methods have been developed to increase the flexibility of $\mathcal{Q}$, e.g. adding auxiliary variables \cite{agakov2004auxiliary,maaloe2016auxiliary} or flow-based methods \cite{challis2012affine,rezende2015variational}, to obtain a tighter ELBO.

Recent works \cite{townsend2019practical,townsend2019hilloc,kingma2019bit} have successfully applied VAE style models to  lossless compression realizing impressive performance. In this setting, the average compression length on the test data set is approximately equal to $-\frac{1}{M}\sum_{m=1}^M \mathrm{ELBO}(x'_m,\theta,\phi)$ (also see Appendix~\ref{app:bbans}). Hence the better the test ELBO indicates the better the compression performance. This motivates us to study the factors that affect the generalization of VAEs and find practical ways to improve the generalization of VAEs.

The contributions of our paper are summarized as follows:
\begin{itemize}
    \item We show the generalization of VAEs is affected by both the generative model (decoder) and the amortized inference network (encoder); and that the overfiting of VAEs is mainly dominated by the amortized inference.
    \item We propose a new training objective that can improve the generalization of the amortized inference without changing the model itself.
    
    \item We demonstrate how the proposed method can improve the compression rate in a practical lossless compression system without scarifying any computation speed. 
\end{itemize}


\section{Generalization of VAEs}\label{sec:generalization_vaes}
During training, we only have access to a finite dataset $\mathcal{X}_{train}$, which leads to the following Monte-Carlo approximation as our objective to train VAEs:
\begin{align}
    \langle\mathrm{ELBO}(x,\theta,\phi)\rangle_{p_d(x)}\approx  \frac{1}{N}\sum_{n=1}^N \mathrm{ELBO}(x_n,\theta,\phi).
\end{align}
\begin{wrapfigure}{r}{0.4\textwidth}
  \vspace{-0.4cm}
  \begin{center}
         \includegraphics[width=\linewidth]{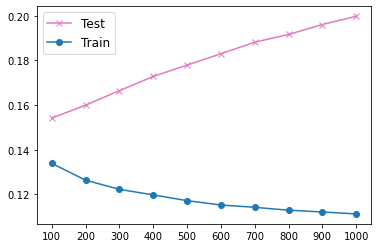}
         \caption{BPD vs epochs. The training BPD decreases but the testing BPD increases during training, which indicates the VAE overfits to $\mathcal{X}_{train}$.
         }
         \label{fig:overfit}
  \end{center}
  \vspace{-0.4cm}
\end{wrapfigure}
This empirical approximation will lead to the VAE overfits to the training data for finite $N$.
 For example,  we train a VAE on the Binary MNIST dataset for 1k epochs and plot the Bits-per-dimension (BPD)\footnote{In the case of VAE, the BPD is defined as the the negative ELBO (with a base 2 logarithm) normalized by the data dimension, lower BPD indicates higher ELBO.} of both training and testing dataset for every 100 epochs, also see
Section 4 for model and training details. Figure \ref{fig:overfit} visualizes the training and testing BPD, which shows the VAE model is overfitting to the training dataset.

The decomposition in Equation \ref{eq:rewrite:ELBO} suggests that the empirical ELBO  contains 1) a \emph{ model empirical approximation}:
\begin{align}
\resizebox{0.5\hsize}{!}{$
    \mathrm{KL}(p_d(x))||p_\theta(x))\approx \frac{1}{N}\sum_{n=1}^N \log p_\theta(x_n)+const.,$}
\end{align}
which will potentially make a flexible model $p_\theta(x)$ overfit to the training data; and 2) an \emph{amortized inference empirical approximation}:
\begin{align}
\resizebox{0.65\hsize}{!}{$%
    \big\langle\mathrm{KL}(q_\phi(z|x)||p_\theta(z|x))\big\rangle_{p_d(x)}\approx \frac{1}{N}\sum_{n=1}^N \mathrm{KL}(q_\phi(z|x_n)||p_\theta(z|x_n))$},\label{eq:inference:approx}
\end{align}
where similarly a flexible $q_\phi(z|x)$ can also overfit to the training data. More specifically, we let $\hat{\phi}$ be the optimal parameter of the empirical variational inference objective
\begin{align}
\resizebox{0.5\hsize}{!}{$%
\hat{\phi}=\arg\min_\phi \frac{1}{N}\sum_{n=1}^N \mathrm{KL}\left(q_\phi(z|x_n)||p_\theta(z|x_n)\right)$}
\end{align}
and we assume for any training data point $x_n\in\mathcal{X}_{train}$ 
\begin{align}
        q_{\hat{\phi}}(z|x_n)&=\arg\min\nolimits_{q\in \mathcal{Q}}\mathrm{KL} (q_{\phi} (z|x_n)||p_\theta(z|x_n))\nonumber \equiv q^*(z|x_n),
\end{align}
where $q^*(z|x_n)$ is the realizable optimal posterior (in the $\mathcal{Q}$ family) for $x_{n}$\footnote{For a powerful  inference network we assume  that there is no amortization gap \citep{cremer2018inference}, which means 
 $q_{\hat{\phi}}(z|x)$ can provide the optimal $q^*(z|x_n)$ for any training data $x_{n}\in \mathcal{X}_{train}$ - see Section \ref{app:related} for further discussion.}. When $q_{\hat{\phi}}(z|x_n)$ overfits to $\mathcal{X}_{train}$,
$q_{\hat{\phi}}(z|x'_m)$ may not be a good approximation to the true posterior $p_\theta(z|x_m')$ for test data $x'_m\in \mathcal{X}_{test}$,
We refer to the difference between the ELBO evaluated using $q_{\hat{\phi}}(z|x)$ and the ELBO evaluated using $q^*(z|x)$ as the \emph{amortized inference generalization gap}, which is formally defined as 
\begin{align}
    \left\langle\mathrm{KL}(q_{\hat{\phi}}(z|x)||p_\theta(z|x)) -\mathrm{KL}(q^*(z|x)||p_\theta(z|x))\right\rangle_{p_d(x)}.
\end{align}
Equivalently, this gap can be written as the difference between two ELBOs with two different $q$
\begin{align}
      \big\langle\underbrace{\langle\log p_\theta(x,z)-\log q^*(z|x)\rangle_{q^*(z|x)}}_{\text{ELBO with optimal inference}}- \underbrace{\langle\log p_\theta(x,z)-\log q_{\hat{\phi}}(z|x)\rangle_{q_{\hat{\phi}}(z|x)}}_{\text{ELBO with amortized inference}}\big\rangle_{p_d(x)}.\label{eq:inference:gap}
\end{align}
The inference neural network introduced by amortization is the cause of this inference generalization gap. It is important to emphasize that this gap cannot be reduced by simply using a more flexible $\mathcal{Q}$. This would only make $\mathrm{KL} (q_{\phi}(z|x_n)||p_\theta(z|x_n))$ smaller for the training data $x_n\in \mathcal{X}_{train}$ but would not explicitly encourage better generalization performance on test data~\cite{shu2019amortized}. 

To summarize, the generalization performance of a VAE depends on two factors:
\begin{itemize}
    \item \textbf{Generative model generalization gap:} defined as $\mathrm{KL}(p_d(x)||p_\theta(x))$ and is caused by  the generative model  overfitting to the the training data.
    \item \textbf{Amortized inference generalization gap:} defined in Equation \ref{eq:inference:gap} and is caused by  the amortized inference model (encoder)  overfitting to the the training data.
\end{itemize}





\subsection{Impact of the Generalization Gaps}
The \emph{generative model generalization gap} that is estimated  by the test dataset (up to a constant) $\mathrm{KL}(p_d(x)||p_\theta(x))\approx  -\frac{1}{M}\sum_{m=1}^{M} \log p_\theta (x'_m)+const.$
cannot be calculated explicitly since we can only evaluate the lower bound $-\frac{1}{M}\sum_{m=1}^{M}\mathrm{ELBO}(x_m',\theta,\phi)$. Fortunately, as suggested in Equation \ref{eq:rewrite:ELBO:0}, if we know the optimal posterior for the test data $q^*(z|x_m')\equiv \arg\min_{q\in\mathcal{Q}}\mathrm{KL}(q(z|x_m')||p_\theta(z|x_m'))$, the log-likelihood can be approximated by the lower bound $\log p_\theta(x_m')\approx \mathrm{ELBO}(x_m',\theta,\phi) $ and the approximation becomes an equality when $p_\theta(z|x_m')\in \mathcal{Q}$. Similarly, 
 the \emph{amortized inference generalization gap} can  be estimated by knowing the optimal posterior $q^*(z|x_m')$ for the test dataset:
\begin{align}
    \frac{1}{M}\sum_{m=1}^M \left\langle\log p_\theta(x_m',z)-\log q^*(z|x_m')\right\rangle_{q^*(z|x_m')}- \langle\log p_\theta(x_m',z)-\log q_{\hat{\phi}}(z|x_m')\rangle_{q_{\hat{\phi}}(z|x_m')}.
\end{align}
We can then estimate $q^*(z|x_m')$ by fixing $\theta$ (which is trained on the training dataset) and learning $\phi^*$ on the test dataset and assuming $q^*(z|x_m')=q_{\phi^*}(z|x_m')$, where
\begin{align}
    \phi^*=\min\nolimits_{\phi}\mathrm{KL}(q_\phi(z|x_m')||p_\theta(z|x_m'))=\max\nolimits_\phi \big\langle\log p_\theta(x_m',z)-\log q_\phi(z|x_m')\big\rangle_{q_\phi(z|x_m')}.\label{eq:optimal:inference}
\end{align}
 This optimal inference strategy can  eliminate the effect of the inference generalization gap, allowing us to isolate the degree to  which both the generative model and amortized inference generalization gaps are contributing to the overfitting.



\begin{wrapfigure}{r}{0.4\textwidth}
\vspace{-0.7cm}
  \begin{center}
    \begin{tikzpicture}
    \node[anchor=south west,inner sep=0] at (0,0) {
        \includegraphics[width=\linewidth]{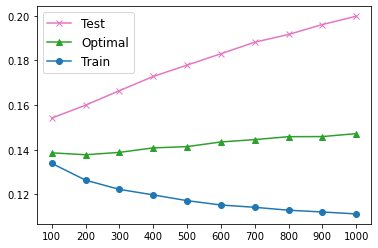}
    };
    \draw[<->,thick, color=red] (2.3,1.55) -- (2.3,2.55) node [pos=0.5,right,align=center, text=red] (TextNode2) {amortized inference \\generalization gap};
\end{tikzpicture}

         \caption{Test BDP vs epochs. Demonstrates the amortized inference generalization gap in a VAE trained on MNIST.}
         \label{fig:inference:generalization:gap}
  \end{center}
  \vspace{-0.4cm}
\end{wrapfigure}

We take the VAE described in Section~\ref{sec:generalization_vaes} and  train $q_\phi(z|x)$ for 1k epochs with Adam~\cite{kingma2014adam} and $\text{lr}=5{\times}10^{-4}$ on the test data using Equation \ref{eq:optimal:inference} to obtain the test BPD for the optimal inference strategy.
In Figure \ref{fig:inference:generalization:gap} we plot the test ELBO (BPD) using the optimal inference strategy (green) and classic amortized inference (purple). Since for the optimal inference strategy the average likelihood $\frac{1}{M}\sum_{n=1}^M \log p_\theta(x'^{(m)})$ can be effectively approximated by the ELBO (see Appendix~\ref{app:tightness} for an empirical verification of the tightness of the ELBO), then the difference between the two inference curves on the test set (Test and Optimal)  is the \emph{amortized inference generalization gap}. We observe that after eliminating the inference generalization gap, the test BPD is stable with a marginal increase during training. This suggests the generative model (decoder) slightly overfit to the data but that the overfitting is mainly dominated by the overfitting of the amortized inference network. 

Although the optimal inference strategy can help eliminate the inference generalization gap, training $q_{\phi}$ on the test data is not practical in most applications of interest.
Therefore, we now focus on improving the generalization of amortized inference without access to the test data at training time.



\section{Improving Generalization with Consistent Amortized Inference}

We now propose an \emph{inference consistency requirement} which, if satisfied, would result in optimal generalization performance for amortized variational inference. Specifically when $p_\theta\rightarrow p_d$, the amortized posterior should converge to the true posterior $q_\phi(z|x)\rightarrow p_\theta(z|x)$\footnote{We assume the true posterior belongs to the variational family $p_\theta(z|x)\in \mathcal{Q}$.} for every $x\sim p_d(x)$. Although this requirement seems natural for variational inference, the classic amortized inference training that is  used for VAEs \cite{kingma2013auto} doesn't satisfy it. Recall the typical VAE empirical ELBO training objective
\begin{align}
\resizebox{0.5\hsize}{!}{$%
\frac{1}{N}\sum_{n=1}^N \log p_\theta(x_n)-\mathrm{KL}(q_\phi(z|x_n)||p_\theta(z|x_n)).$}
\end{align}
When the model converges to the true distribution $p_{\theta^*}=p_d$ the training criterion for $q_\phi(z|x)$
\begin{align}
\resizebox{0.47\hsize}{!}{$%
   \min\nolimits_\phi-\frac{1}{N}\sum_{n=1}^N \mathrm{KL}(q_\phi(z|x_n)||p_{\theta^*}(z|x_n))$}
\end{align}
can still result in the amortized posterior $q_\phi(z|x)$ overfitting to the training data. In principle, one could  also limit the network capacity and/or add an explicit regularizer to the parameters \cite{shalev2014understanding} in an attempt to improve the generalization. However, this still cannot satisfy the consistency requirement in principle because it still only use  the finite training dataset. Alternatively, there is another classic variational inference method that we now discuss, the wake-sleep training algorithm~\cite{dayan1995helmholtz,hinton1995wake}, which does in fact satisfy the proposed consistency requirement.

\subsection{Wake-Sleep Training}
 Defining $q_\phi(x,z)=q_\phi(z|x)p_d(x)$ and $p_\theta(x,z)=p_\theta(x|z)p(z)$, the two phases of the wake-sleep training \cite{dayan1995helmholtz,hinton1995wake} can be written as minimizing two different KL divergences in both $x$ and $z$ space.

\textbf{Wake phase model learning:} $p_\theta(x|z)$ is trained by minimizing the KL divergence
\begin{align}
    &\min\nolimits_\theta \mathrm{KL}(q_\phi(x,z)||p_\theta(x,z))=  \max\nolimits_\theta\big\langle \mathrm{ELBO}(x,\theta,\phi)\big\rangle_{p_d(x)}+const.,
\end{align}
where $\langle\cdot\rangle_{p_d(x)}$ is approximated  using the training set.
This is referred to as the \emph{wake phase} since the model is trained on experience from the `real environment', i.e. it uses true data samples from $p_d(x)$. 

\textbf{Sleep phase amortized inference:}  $q_\phi(z|x)$ is trained by minimizing the KL divergence
\begin{align}
&\min\nolimits_\phi \mathrm{KL}(p_\theta(x,z)||q_\phi(x,z))=\min\nolimits_\phi \big\langle\mathrm{KL} (p_\theta(z|x)||q_\phi(z|x))\big\rangle_{p_\theta(x)}+const.\label{eq:sleep}
\end{align}
Leaving out the terms that are irrelevant to $\phi$, the objective can be estimated with Monte-Carlo
$\langle-\log q_\phi(z|x_m')\rangle_{p_\theta(x,z)}\approx \frac{1}{K} \sum_{k=1}^K -\log q_\phi(z_k|x_k),$
where $z_k\sim p(z)$ and $x_k\sim p_\theta(x|z_k)$. This is referred to as the \emph{sleep phase} because the samples from the model used to train $q_\phi$ are interpreted as dreamed experience. In contrast, the training criterion for the typical VAE amortized inference (Equation \ref{eq:inference:approx}) uses the true data samples from $p_d$ to train $q_\phi(z|x)$, which we  refer to as \emph{wake phase amortized inference}. 
We notice that if a perfect model $p_{\theta^*}(x)=p_d(x)$ is used in the sleep phase amortized inference, then it is equivalent to  minimizing
\begin{align}
     \big\langle\mathrm{KL}(p_\theta(z|x)||q_\phi(z|x))\big\rangle_{p_{\theta^*}(x)}= \big\langle\mathrm{KL}(p_\theta(z|x)||q_\phi(z|x))\big\rangle_{p_{d}(x)}.
\end{align}
Therefore, the training of the inference network satisfies the \emph{inference consistency requirement} since we can access infinite training data from $p_d$ by sampling from $p_{\theta^*}$.

However, the wake-sleep algorithm presented lacks convergence guarantees~\cite{dayan1995helmholtz} and minimizing $\mathrm{KL}(p_\theta(z|x)||q_\phi(z|x))$ in the sleep phase doesn't necessarily encourage an improvement to the ELBO, which directly relates to the compression rate in the lossless compression application \cite{townsend2019practical}. Therefore, in the next section, we propose a new variational inference scheme: \emph{reverse sleep amortized inference} and demonstrate how it helps improve the generalization of the inference network in practice.

\subsection{Reverse Sleep Amortized Inference}
We propose to use the \emph{reverse KL divergence} in the sleep phase. We  fix $\theta$ and train $\phi$ using
\begin{align}
     \min\nolimits_\phi\big\langle\mathrm{KL}(q_\phi(z|x)||p_\theta(z|x)\big\rangle_{p_\theta(x)}&=\max\nolimits_\phi\big\langle \log p_\theta(x,z)-\log q_\phi(z|x)\big\rangle_{q_\phi(z|x)p_\theta(x)},\label{eq:reverse:sleep}
\end{align}
where the integration $\langle\cdot\rangle_{p_\theta(x)}$ is approximated by Monte-Carlo using samples from the generative model $p_\theta(x)$. This reverse KL objective encourages improvements to the ELBO.  When we have a perfect model $p_{\theta^*}(x)=p_d(x)$ the reverse sleep phase is equivalent to 
\begin{align}
\min\nolimits_\phi \big\langle\mathrm{KL}(p_{\theta^*}(z|x)||q_\phi(z|x))\big\rangle_{p_{\theta^*}(x)}&=\min\nolimits_\phi \big\langle\mathrm{KL}(p_{\theta^*}(z|x)||q_\phi(z|x))\big\rangle_{p_{d}(x)}
\end{align}
which satisfies the \emph{inference consistency requirement}.

\begin{wrapfigure}{r}{0.4\textwidth}
    \vspace{-1cm}
         \centering
        \includegraphics[width=0.4\textwidth]{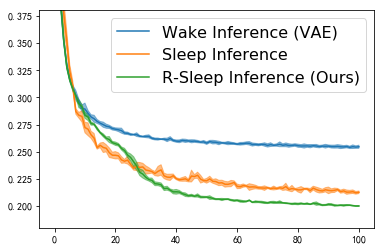}
            \caption{Test BPD vs epochs. We compare the consistency property between three amortized inference methods. \label{fig:true:model}}
    \vspace{-0.2cm}
\end{wrapfigure}
The consistency requirement can also be validated empirically when the perfect model is known $p_{\theta^*}(x)=p_d(x)$. This can be achieved by using a pre-trained VAE as the true data generation distribution. Therefore, we first train a VAE to fit the binary MNIST problem. The VAE has the same structure as that used in Section 2 and is trained using Adam with $lr=1{\times}10^{-3}$ for 100 epochs. After training, we treat the pre-trained decoder $p_{\theta'}(x|z)$ as the training data generator $p_d(x)\equiv \int p_{\theta'}(x|z)p(z)dz$.  We then sample 10000 data samples from $p_{d}$ to form a training set $\mathcal{X}_{train}$ and 1000 samples to form a test set $\mathcal{X}_{test}$. 
We then train a new $q_\phi(z|x)$ with: 1) wake phase  inference (VAE) 2) (forward) sleep  inference and 3) reverse sleep inference. The network is trained using Adam with $lr=1{\times}10^{-3}$ for 100 epochs.
Figure \ref{fig:true:model} shows the test BPD calculated after every training epoch. We can see the sleep phase out-performs the wake phase and the reverse sleep inference achieves the best BPD. Intuitively, this is because both the forward and reverse sleep inference use the true model to generate additional training data whereas the wake inference only has access to the finite training dataset $\mathrm{X}_{train}$.

\subsection{Reverse Half-asleep Amortized Inference with Imperfect Models}
In practice our model will not be perfect $p_\theta\neq p_d$. Empirically we find that samples from even a well trained model $p_\theta$ may not always be sufficiently like the samples from the true data distribution. This can lead to degradation in the performance of the inference network when using the reverse-sleep approach. For this reason,  we propose to use a mixture distribution between the model and the empirical training data distribution as follows
\begin{align}
\label{eq:inference:approx:mx}
\big\langle\mathrm{KL}\big(q_\phi(z|x)||p_\theta(z|x)\big)\big\rangle_{m(x)} \quad \text{where}\quad  m(x)\equiv\alpha p_\theta(x)+(1-\alpha)\hat{p}_d.
\end{align}
When $\alpha=0$, it reduces to the standard approach used in VAE training. 
When $\alpha=1$, we recover the reverse sleep method (Equation \ref{eq:reverse:sleep}). We find that a setting of $\alpha=0.5$ works well in practice. This balances samples from the true underlying data distribution with samples from the model. 

We thus refer to this method as \emph{reverse half-asleep} since it uses both data  and model samples to train the amortized posterior. Intuitively, we can rewrite the Equation \ref{eq:inference:approx:mx} as a sum of two positive terms
\begin{align}
\alpha\big\langle\mathrm{KL}\big(q_\phi(z|x)||p_\theta(z|x)\big)\big\rangle_{\hat{p}(x)}+(1-\alpha)\big\langle\mathrm{KL}\big(q_\phi(z|x)||p_\theta(z|x)\big)\big\rangle_{p_\theta(x)}.
\end{align}
Therefore, the optimal of this objective will make the first term 0, which is the same requirement as the classic amortized inference (Equation \ref{eq:inference:approx}). The second term, which is equivalent  to the 
\begin{wrapfigure}{r}{0.4\textwidth}
         \centering
        \includegraphics[width=0.4\textwidth]{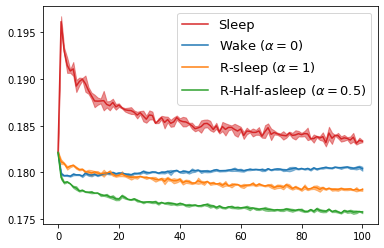}
        \caption{Test BPD comparisons of Amortized inference with different $\alpha$. We find the Reverse Half-asleep method ($\alpha=0.5$) achieves the best BPD. The mean and std are calculated with three random seeds. }
        \label{fig:learned:model}
        \vspace{-0.2cm}
\end{wrapfigure}
reverse sleep amortized inference (Equation \ref{eq:reverse:sleep}),
 can encourage the inference consistency requirement: when $p_\theta=p_d$, the optimal of the second term will set $q_\phi(z|x)=p_\theta(z|x)$ for any $x\sim p_d(x)$. When  $p_\theta$ is not perfect, 
 the second term can be seen as a regularizer added to the classic amortized inference objective, which can be used to penalize the hypothesis space of the amortized network~\cite{shalev2014understanding}.
 
To compare with different $\alpha$, we first fit a VAE (with the same structure as that used in Figure 2)  to the Binary MNIST dataset, and then train the amortized posterior using sleep inference (Equation \ref{eq:sleep}) and three different $\alpha$  for additional 100 epochs using Adam with learning rate $3{\times}10^{-4}$.
Figure \ref{fig:learned:model} shows the test BPD comparison. We find the proposed reverse half-asleep method ($\alpha=0.5$) outperforms the reversed sleep method ($\alpha=1$), whereas the standard amortized inference training in VAE ($\alpha=0$) leads to overfitting of the inference network. We also plot the sleep inference training curve, whose BPD is less competitive since it is not directly optimizing the ELBO.

\section{Generalization Experiments}
\label{sec:experiments}
\begin{figure*}[t]
     \centering
     \begin{subfigure}[b]{0.32\textwidth}
         \centering
         \includegraphics[width=\textwidth]{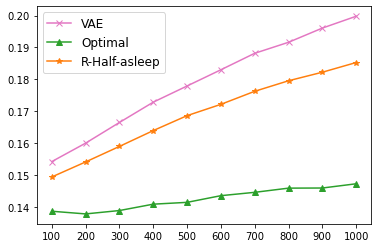}
         \caption{Binary MNIST}
         \label{fig:binary}
     \end{subfigure}
     \begin{subfigure}[b]{0.32\textwidth}
         \centering
         \includegraphics[width=\textwidth]{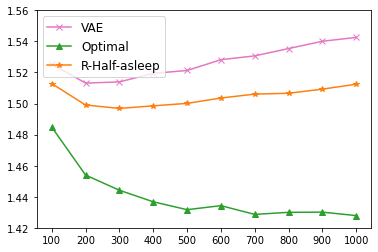}
         \caption{Grey MNIST}
            \label{fig:grey}

     \end{subfigure}
      \begin{subfigure}[b]{0.32\textwidth}
         \centering
         \includegraphics[width=\textwidth]{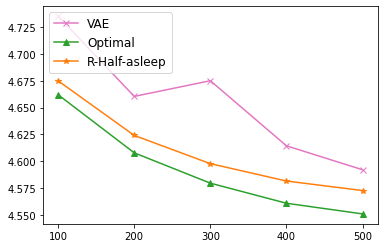}
         \caption{CIFAR10}
         \label{fig:cifar}
     \end{subfigure}
        \caption{Test BPD comparisons among amortized inference (VAE), optimal inference strategy and the  reverse half-asleep inference on three datasets. The x-axis represents the training epochs.}
        \label{fig:three graphs}
        \vspace{-0.3cm}
\end{figure*}

 We  apply the  reverse half-asleep  to improve the generalization of VAEs on three different datasets: binary MNIST, grey MNIST~\cite{lecun1998mnist} and CIFAR10~\cite{krizhevsky2009learning}. For binary and grey MNIST, we use latent dimension 16/32 and neural nets with 2 layers of 500 hidden units in both the encoder and decoder. We use Bernoulli $p(x|z)$ for binary MNIST and discretized logistic distribution for grey MNIST. We  train the VAE with the usual amortized inference approach using Adam with $lr=3{\times} 10^{-4}$ for 1000 epochs and save the model every 100 epochs. We then use the saved models to 1) evaluate on the test data sets, 2) conduct optimal inference by training $q_\phi(z|x)$ on the test data and 3) run reverse half-asleep method before calculating the test BPD. For the reverse half-asleep, we train the amortized posterior for 100 epochs with Adam and $lr=5{\times}10^{-4}$. To sample from $p_\theta(x)$, we firstly sample $z'\sim p(z)$ and sample $x'\sim p(x|z=z')$. For the optimal inference strategy, we train the amortized posterior with the same optimization scheme on the test data set for additional 500 epochs to ensure the same number of gradient steps are conducted (since training set is 5 times as big as the test set). Figure \ref{fig:binary} and \ref{fig:grey} show the test BPD comparisons of binary and grey MNIST respectively and demostrate that our approach does not require further training on the test data to improve generalization performance. 

 For CIFAR10, we use the convolutional ResNet \cite{he2016deep,van2017neural} with 2 residual blocks and latent size 128.  The observational distribution is a discretized logistic distribution with linear autoregressive parameterization within channels.
 We train the VAE for 500 epochs with Adam and $lr=5{\times}10^{-4}$ and save the model every 100 epoch. The pre-trained VAE achieves 4.592 BPD on the CIFAR10, which is comparable with other single latent VAE models reported in \cite{van2017neural}: 4.51 BPD with a VAE with latent dimension 256 and 4.67 BPD with a discrete latent VAE (VQVAE).
 
 Ideally, when the VAE model converges to the true distribution $p_\theta\rightarrow p_d$, the aggregate posterior $q_\phi(z)=\int q_\phi(z|x)p_d(x)dx$ will match the prior $p(z)$.
However, for a complex distribution like CIFAR10,  a significant mismatch between $q_\phi(z)$ and $p(z)$ is usually observed in practice \cite{zhao2017infovae,dai2019diagnosing}. In this case, the sample $x'$ that is generated using a latent sample from the prior $x'\sim p_\theta(x|z')$, where $z'\sim p(z)$, may be blurry or invalid.
A common solution is to  train another model, e.g. a VAE  \cite{dai2019diagnosing} or a PixelCNN \cite{van2016pixel,van2017neural} to approximate $q_\phi(z)$. In our case, we instead directly sample from $q_\phi(z)$ rather than $p(z)$ to generate samples in Equation \ref{eq:reverse:sleep}, which can be done by first sampling $x'\sim p_d(x)$ (from the training dataset) and then sample $z'\sim q_\phi(z|x=x')$. This scheme still results in a consistent training objective since $q_{\phi^*}(z)=p(z)$ for the optimal posterior $q_{\phi^*}(z|x)$. We use Adam with $lr=1{\times}10^{-5}$ and train the reverse half-asleep inference  for 100 epochs on the training data and train the optimal inference strategy for 500 epochs on the test data, see Figure \ref{fig:cifar} for the result. 
  We find the proposed reverse half-asleep training  approach (with sampling from $q_\phi(z)$) consistently improves the generalization performance of the amortized posterior. We also apply the proposed method on a VAE trained on CIFAR100 for 500 epochs (the rest of the experiment settings are the same as the CIFAR10 case) and  find our method improves the BPD from 5.288 to 5.275.
 
\subsection{Comparisons with Regularization Methods}
\begin{figure}[h]
     \begin{subfigure}[b]{0.32\textwidth}
         \centering
         \includegraphics[width=\textwidth]{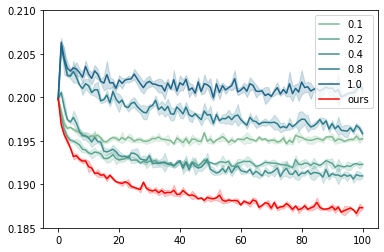}
         \caption{Binary MNIST}
     \end{subfigure}
     \begin{subfigure}[b]{0.32\textwidth}
         \centering
     \includegraphics[width=\textwidth]{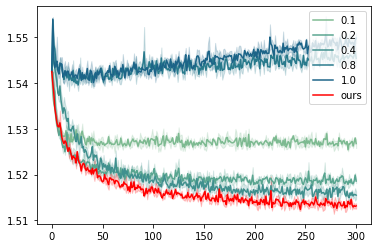}
     \caption{Grey MNIST}
     \end{subfigure}
      \begin{subfigure}[b]{0.32\textwidth}
         \centering
     \includegraphics[width=\textwidth]{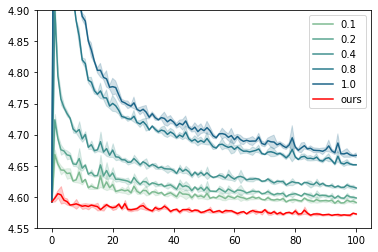}
     \caption{CIFAR 10}
     \end{subfigure}
     \caption{Test BPD evaluated after every training epoch. We find, compared to the denoising regularizer, the proposed amortized inference training scheme consistently achieves better generalization performance in all tasks.\label{fig:comparison}}
\end{figure}

Recent work ~\cite{shu2019amortized} proposed to alleviate overfitting of amortized inference by optimizing a linear combination between the traditional amortized inference (Equation \ref{eq:inference:approx}) and a denoising objective
 \begin{align}
     \alpha\left\langle\mathrm{KL}(q_\phi(z|x+\epsilon)||p_\theta(z|x))\right\rangle_{p(\epsilon)}+(1-\alpha) \mathrm{KL}(q_\phi(z|x)||p_\theta(z|x)),
 \end{align}
where $p(\epsilon)= \mathcal{N}(0,\sigma^2I)$. We compare this regularizer to our method by training the amortized posterior of VAEs for an additional 100, 300 and 100 epochs on Binary, Grey MNSIT and CIFAR respectively. For the denoising regularizer, we use the same linear combination weight $\alpha=0.5$ as that used in Equation \ref{eq:inference:approx:mx} and vary $\sigma \in \{0.1,0.2,0.4,0.6,0.8,1.0\}$, see  Table \ref{tab:comparison} for the  comparisons. For MNIST, we find $\sigma\in\{0.1,0.2,0.4\}$ improves the generalization but larger noise levels hurts the performance. For CIFAR10, only $\sigma=0.1$ can slightly improve the generalization by 0.001 BPD.  In contrast, our method consistently achieves better generalization performance without tuning any hyper-parameters, see
Figure \ref{fig:comparison} for the test BPD (evaluated every training epoch, the mean/std are calculated with 3 random seeds). Compared to the denoising approach, one limitation of our method is the requirement of model samples, which is more computational expansive during training.

Since the decoder is shared and fixed in all comparisons, better test ELBO indicates the predicted $q_\phi(z|x')$ is closer to the true posterior $p_\theta(z|x')$ under the KL divergence measure (see Equation \ref{eq:rewrite:ELBO:0}, higher ELBO with fixed $\theta$ indicates $\mathrm{KL}(q_\phi(z|x)||p_\theta(z|x))$ is smaller). Therefore, the proposed method can also benefit a range of tasks that require accurate prediction of the posterior on the test data. In Appendix \ref{app:tightness} and \ref{app:representation}, we demonstrate our method can provide better proposal distributions for the importance weighted Auto-Encoder~\cite{burda2015importance} and also improve the representation learning performance for down-stream classification tasks.
\begin{table}[h]
    \centering
\caption{Average test BPD comparisons with  Denoising Regularizer~\cite{shu2019amortized}.\label{tab:comparison}}
\begin{tabular}{ cccccccc } 
\hline
Methods & VAE & $\sigma=0.1$ &$\sigma=0.2$ & $\sigma=0.4$ & $\sigma=0.8$& $\sigma=1.0$ & Ours \\ 
 \hline
Binary MNIST& 0.200& 0.195& 0.192 & 0.191 & 0.196 & 0.201  & \textbf{0.187}\\
 Grey MNIST& 1.543& 1.527& 1.519  & 1.515 & 1.545 &1.550  & \textbf{1.513}\\ 
CIFAR10 & 4.592& 4.591& 4.598 & 4.614 & 4.651 & 4.667  & \textbf{4.572}\\ 
 \hline
\end{tabular}
\end{table}


\section{Application of Lossless Compression}
Lossless compression is an important application of VAEs where generalization plays a key role in the compression rate. Given a VAE with $p_\theta(x|z)$, $q_\phi(z|x)$ and $p(z)$, a practical compressor can be efficiently implemented using the Bits Back algorithm \cite{hinton1993keeping,townsend2019practical} with the ANS coder~\cite{duda2013asymmetric}. See Appendix~\ref{app:bbans} for a detailed introduction of conducting lossless compression with VAE models.
In  Algorithm \ref{algo:bbAIVAE}, we summarize the Bits Back procedure with amortized inference to compress/decompress a test data point $x'$ to a stack that contains bit string messages.
The resulting code length for data $x'$ is approximately equal to the negative ELBO
\begin{align}
    -\log_2 p_\theta(x'|z')- \log_2 p(z')+\log_2 q_\phi(z'|x').
\end{align}

\begin{wrapfigure}{r}{0.55\textwidth}
    \begin{minipage}{0.55\textwidth}
    \vspace{-0.8cm}
    \begin{algorithm}[H]
\captionof{algorithm}{Bits Back with Amortized Inference.\label{algo:bbAIVAE}}
Comp./decomp. stages share $\{p_\theta(x|z),q_\phi(z|x),p(z)\}$.
\ruleline{\color{JungleGreen}{Compression}}

  \begin{algorithmic}
    \STATE  Draw sample $z'\sim q_\phi(z|x')$ from the stack.
    \STATE   Encode $x'\sim p_\theta(x|z')$ onto the stack.
    \STATE   Encode  $z'\sim p(z)$ onto the stack.
  \end{algorithmic}
\ruleline{\color{JungleGreen}{Decompression}}

\begin{algorithmic}
\STATE  Decode $z'\sim p(z)$  from the stack.
\STATE  Decode $x'\sim p_\theta(x|z')$ from the stack.
\STATE  Encode $z'\sim q_\phi(z|x')$ onto the stack.
  \end{algorithmic}
  \end{algorithm}
\end{minipage}
\end{wrapfigure}
We have shown that  $q_\phi(z|x)$ may overfit to the training data, degrading compression performance. To improve the compression BPD, the optimal inference strategy can also be applied in the Bits Back algorithm. In the compression stage, we can train $\phi$ by
\begin{align}
\phi^*=\arg\max\nolimits_\phi \mathrm{ELBO}(x',\theta,\phi)\label{eq:encoder:train}.
\end{align}
When the $q_\phi(z|x')$ is parameterized to be a Gaussian, we can just take $\phi$ to be the mean and standard deviation $\mathcal{N}(\phi_{\mu},\phi_{\sigma}^2)$, which only contains two training parameters.
In the decompression stage, we observe that the compressed data $x'$ is recovered before the $q_\phi(z|x')$ is used to encode $z'$.
Therefore, we can also train the $q_\phi(z|x')$ using the recovered  $x'$ to maximize the test ELBO. If the optimization procedure is the same as that used in the compression stage, we will get the same $q_{\phi^*}(z|x')$.  In practice, we need to pre-specify the number of gradient descent steps $K$. When $K$ is large, we recover the optimal inference strategy and the code length is approximately
\begin{align}
     -\log_2 p_\theta(x'|z')- \log_2 p(z')+\log_2 q_{\phi^*}(z'|x').\label{eq:bboi:length}
\end{align}
\begin{wrapfigure}{R}{0.55\textwidth}
    \begin{minipage}{0.55\textwidth}
    \vspace{-0.3cm}
 \begin{algorithm}[H]
\caption{Bits Back with $K$-step Optimal Inference\label{algo:bbOIVAE}}
Comp./decomp. stages share $\{p_\theta(x|z),q_\phi(z|x),p(z)\}$ and the optimization procedure of Equation \ref{eq:encoder:train}.
\ruleline{\color{JungleGreen}{Compression}}

  \begin{algorithmic}
  \STATE Take K gradient steps $\phi\rightarrow\phi^K$ with Equation   \ref{eq:encoder:train}.
    \STATE  Draw sample $z'\sim q_{\phi^K}(z|x')$ from the stack.
   \STATE   Encode $x'\sim p_\theta(x|z')$ onto the stack.
    \STATE  Encode $z'\sim p(z)$ onto the stack.
  \end{algorithmic}
\ruleline{\color{JungleGreen}{Decompression}}

  \begin{algorithmic}
\STATE  Decode $z'\sim p(z)$  from the stack.
\STATE  Decode $x'\sim p_\theta(x|z')$ from the stack.
  \STATE Take K gradient steps $\phi\rightarrow\phi^K$ with Equation   \ref{eq:encoder:train}.
\STATE  Encode $z'\sim q_{\phi^K}(z|x')$ onto the stack.
  \end{algorithmic} 
\end{algorithm}
\end{minipage}
\end{wrapfigure}
This observation was first proposed in \cite{yang2020improving} in the context of lossy  compression and then applied to lossless compression with Bits Back coding in \cite{ruan2021improving}. Furthermore, by varying the optimization steps $K$ in the optimal inference, we can trade off between the speed and the compression rate. 
This is valuable for practical applications with different speed/rate requirements. See 
Algorithm \ref{algo:bbOIVAE} for a summary of the Bits Back algorithm with $K$-step optimal inference.

Although the optimal inference strategy can  be used in lossless compression,  it requires extra run-time for training at the compression stages. In contrast, our proposed reverse half-asleep inference scheme can improve the compression rate  \emph{without} scarifying any speed.
Additionally, our method can also provide a better initialization for the optimal-inference strategy to allow a better trade-off between compression rate and speed.

We implement\footnote{Implementation can be found in the following repo: \url{https://github.com/zmtomorrow/GeneralizationGapInAmortizedInference}. All  experiments are run on a NVIDIA V100 GPU.} Bits Back with ANS~\cite{duda2013asymmetric} and compare the compression among four inference methods:

 \textbf{1. Baseline:} This is the classic VAE-based compression introduced by \cite{townsend2019practical}. For binary and grey MNIST, both the encoder and decoder contain 2 fully connected layers with 500 hidden units and latent dimension 10. The observation distributions are Bernoulli and discretized Logistic distribution respectively. For CIFAR10, we use fully convolutional ResNets \cite{he2016deep} with 3 residual blocks in the encoder/decoder, latent dimension 128 and discreteized Logistic distribution with channel-wise linear autoregressive\cite{salimans2017pixelcnn++} as the observation distribution. We train both the amortized posterior and the decoder by maximizing the ELBO (Equation \ref{eq:elbo}) using Adam with $lr=3{\times}10^{-4}$ for 100, 100 and 500 epochs (for Binary MNIST, Grey MNIST and CIFAR10 respectively), and then apply Algorithm \ref{algo:bbAIVAE} to conduct compression.
    
\textbf{2. Reversed Half-asleep}: we do amortized inference using Equation \ref{eq:inference:approx:mx} for 100 and 300 epochs with Adam optimizer ($lr=3{\times}10^{-4}$) for binary and grey MNIST respectively, and $lr=1{\times}10^{-5}$ for 100 epochs for CIFAR10. Other training details are the same as the baseline method.
    
\textbf{3. Optimal Inference:} we take the amortized posterior (encoder)  and decoder from the baseline and apply the $K$-step optimal inference strategy described in Algorithm \ref{algo:bbOIVAE} to do compression. We use Adam optimizer and vary the $K$ from 1 to 10 to achieve a trade-off curve between compression rate and speed. We actively choose the highest learning rate that can make the BPD consistently improve with the increment of $K$: $lr=5{\times}10^{-3}$ for binary and grey MNIST and $lr=1{\times}10^{-3}$ for CIFAR10.
    
\textbf{4. Reversed Half-asleep + Optimal Inference:} we take the encoder in method 2 and  decoder from the baseline and conduct $K$-step optimal inference. All other training details are as per method 3.
\begin{figure*}[h]
     \centering
     \begin{subfigure}[b]{0.32\textwidth}
         \centering
         \includegraphics[width=\textwidth]{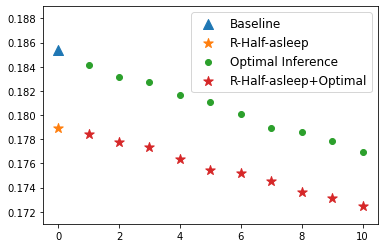}
         \caption{Binary MNIST}
     \end{subfigure}
     \begin{subfigure}[b]{0.32\textwidth}
         \centering
        \includegraphics[width=\textwidth]{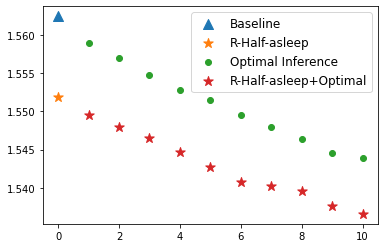}
        \caption{Grey MNIST}
     \end{subfigure}
     \begin{subfigure}[b]{0.32\textwidth}
         \centering
        \includegraphics[width=\textwidth]{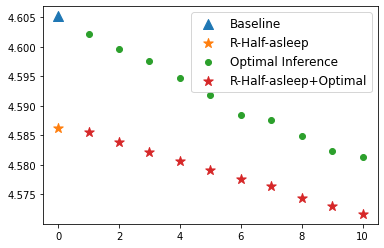}
        \caption{CIFAR10}
     \end{subfigure}
        \caption{We plot the comparisons for different methods. The y-axis is the BPD  and  x-axis represents the $K$ gradient steps in the optimal inference. The baseline and our R-Half-sleep can be seen as  special cases of optimal inference with $K=0$. We find  given a fixed computational budget, our  method achieves a lower BPD than one using traditional amortized inference training. \label{fig:trade:off}}
        
\end{figure*}
\begin{figure}[h]
    \centering
    \small
    \begin{subfigure}{.48\textwidth}
    \centering
        \begin{tabular}{c|ccc}
     \hline
    &Baseline & Ours & K=7
    \\
     \hline
    BPD & 0.185 &  0.179 & 0.179\\
    Com. Time  & 0.006 & 0.006& 0.013\\ 
    Dec. Time & 0.006 & 0.006 & 0.013\\
     \hline
     Time Cost & - & 0\% & \textcolor{red}{116.7\%}  \\
    \hline
    \end{tabular}
    \caption{MNIST}
    \end{subfigure}
    \begin{subfigure}{.48\textwidth}
        \centering
    \small
        \begin{tabular}{c|ccc}
     \hline
    &Baseline & Ours & K=8
    \\
     \hline
    BPD & 4.602 &  4.585 & 4.585\\
    Com. Time & 0.27 & 0.27& 0.38\\ 
    Dec. Time & 0.26 & 0.26 & 0.38\\
    \hline
     Time Cost & - & 0\% & \textcolor{red}{46.2\%}  \\
    \hline
    \end{tabular}
        \caption{CIFAR10}
    \end{subfigure}
    \caption{Compression (Com.) and decompression (Dec.) time comparison. We show that to achieve the same BPD as our method, the $K$-step optimal inference strategy that initializes the amortized posterior needs $K=7$ (binary MNIST) and $K=8$ (CIFAR10) steps for each test datapoint, which will cost an additional $116.7\%$ and $46.2\%$ of time respectively during compression.}
    \label{tab:BPD:time}
\end{figure}


In Figure \ref{fig:trade:off}, we plot test BPD comparisons for the different methods outlined.  We can see if optimization is not allowed at compression time, the use of our reverse-half-asleep method achieves better compression rate with no additional computational cost. If we allow $K$-step optimization during compression, for a given computational budget, the amortized posterior initialized using our reverse-half-asleep method also achieves lower BPD, which leads to a better trade-off between the time and compression rate. Table \ref{tab:BPD:time} also reports the average time improvements of our method to compress a single MNIST and CIFAR10 image respectively, which shows the effectiveness of our method.

\section{Related Work\label{app:related}}
A different perspective on generative models' generalization is proposed in paper \cite{zhao2018bias} where the generalization is evaluated by testing if the model can generate novel combinations of features. However, the  generalization defined in our work is purely measured by the test likelihood, which is a different perspective and more relevant for the application of lossless compression.

Recent work \cite{zhang2021ood} first studies the likelihood-based generalization for lossless compression. They focus on the test and train data that are from different distributions whereas we assume they follow the same distribution. Additionally, their model has a tractable likelihood and relates to the \emph{generative model related generalization}, whereas we focus on \emph{inference related generalization} in VAEs.

Previous work \citep{cremer2018inference} studied the \emph{amortization gap} in amortized inference, which  is caused by using $q_{\phi^*}(z|x_n)$ to generate posteriors for each input $x_n$ rather than learning a posterior $q_n^*(z)$ for  $x_n$ individually. This gap can be alleviated using a larger capacity encoder network.  This amortization gap is fundamentally different from the \emph{inference generalization gap} we discuss in this work since the latter focuses solely on test time generalization but the former problem also exists at training time.

Recent work \cite{ruan2021improving} proposes a compression scheme based on the IWAE~\cite{burda2015importance} bound, which is tighter than the ELBO and thus improves the compression rate. However, this method has to compress/decompress multiple latent samples, which requires extra time cost. On the other hand, we focus on improving the ELBO-based compression that only needs to compress one single latent sample. Nevertheless, similar to the $K$-step optimal inference strategy, our amortized training objective can also be used in the IWAE-based method, which gives a better proposal distribution for importance sampling, see Appendix~\ref{app:tightness} for a demonstration.

Paper \cite{cemgil2020autoencoding} considers the following data generation procedure $x_1\sim p_d(x)$, $z_1\sim p_\theta(z|x_1)$, $x_2\sim p_\theta(x|z_1)$ and propose to enforce latent consistency between $q_\phi(z|x_1)$ and $q_\phi(z|x_2)$ for paired data $(x_1,x_2)$ to encourage the robustness of the learned representation. This procedure is close to the self-supervised contrasting learning method~\cite{chen2020simple} where the augmented data is the reconstruction of the training data using the VAE model. In our paper, we want to encourage the sample from the model $x'\sim \int p_\theta(x|z)p(z)dz$ to have high ELBO under the model (Equation \ref{eq:reverse:sleep}) to improve the generalization of the amortized inference and no paired data is required in our procedure. Therefore, both motivations and methodologies are different from our method.

\section{Conclusion}
We have shown how the generalization of VAEs is largely affected by the amortized inference network and proposed a new variational inference scheme that provides better generalization as demonstrated in the application of lossless compression. Future work will study the generalization of the decoder model to further improve the  performance of VAEs.

%

\clearpage
\bibliography{ref}
\bibliographystyle{abbrv}

\newpage
\section*{Checklist}


\begin{enumerate}

\item For all authors...
\begin{enumerate}
  \item Do the main claims made in the abstract and introduction accurately reflect the paper's contributions and scope?
    \answerYes{}
  \item Did you describe the limitations of your work?
    \answerYes{In Section 4.}
  \item Did you discuss any potential negative societal impacts of your work?
   \answerNA{}
  \item Have you read the ethics review guidelines and ensured that your paper conforms to them?
    \answerYes{}
\end{enumerate}

\item If you are including theoretical results...
\begin{enumerate}
  \item Did you state the full set of assumptions of all theoretical results?
    \answerNA{}
        \item Did you include complete proofs of all theoretical results?
    \answerNA{}
\end{enumerate}

\item If you ran experiments...
\begin{enumerate}
  \item Did you include the code, data, and instructions needed to reproduce the main experimental results (either in the supplemental material or as a URL)?
   \answerYes{See the footnote in  page 8.}
  \item Did you specify all the training details (e.g., data splits, hyperparameters, how they were chosen)?
    \answerYes{See section 4 and 5.}
        \item Did you report error bars (e.g., with respect to the random seed after running experiments multiple times)?
    \answerYes{See Figure 6.}
        \item Did you include the total amount of compute and the type of resources used (e.g., type of GPUs, internal cluster, or cloud provider)?
     \answerYes{See the footnote in  page 8.}
\end{enumerate}

\item If you are using existing assets (e.g., code, data, models) or curating/releasing new assets...
\begin{enumerate}
  \item If your work uses existing assets, did you cite the creators?
    \answerYes{We cite the datasets in section 4.}
  \item Did you mention the license of the assets?
    \answerNA{We didn't find licences for MNIST and CIFAR10.}
  \item Did you include any new assets either in the supplemental material or as a URL?
    \answerNA{}
  \item Did you discuss whether and how consent was obtained from people whose data you're using/curating?
    \answerNA{}
  \item Did you discuss whether the data you are using/curating contains personally identifiable information or offensive content?
    \answerNA{}
\end{enumerate}

\item If you used crowdsourcing or conducted research with human subjects...
\begin{enumerate}
  \item Did you include the full text of instructions given to participants and screenshots, if applicable?
    \answerNA{}
  \item Did you describe any potential participant risks, with links to Institutional Review Board (IRB) approvals, if applicable?
    \answerNA{}
  \item Did you include the estimated hourly wage paid to participants and the total amount spent on participant compensation?
    \answerNA{}
\end{enumerate}

\end{enumerate}

\newpage

\appendix
\section{Tightness of the ELBO and IWAE Improvement \label{app:tightness}}
\begin{figure*}[h]
     \centering
     \begin{subfigure}[b]{0.32\textwidth}
         \centering
         \includegraphics[width=\textwidth]{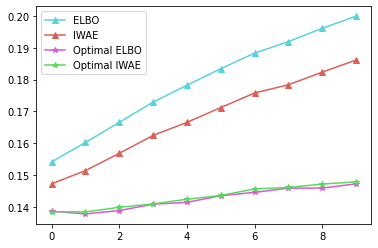}
         \caption{Tightness}
     \end{subfigure}
      \begin{subfigure}[b]{0.32\textwidth}
         \centering
         \includegraphics[width=\textwidth]{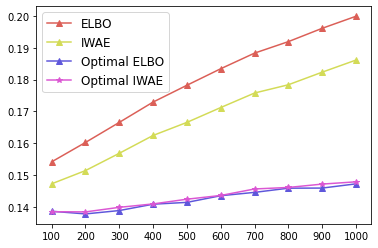}
         \caption{IWAE Improvement}
     \end{subfigure}
     \begin{subfigure}[b]{0.32\textwidth}
         \centering
         \includegraphics[width=\textwidth]{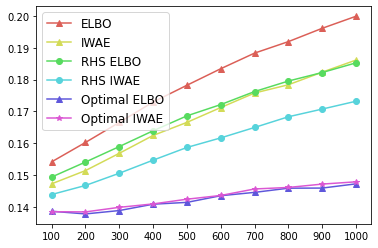}
         \caption{All Comparisons}
     \end{subfigure}
        \caption{IWAE comparisons on Binary MNIST. The x-axis indicates the training epoch and the y-axis is the Bits-per-dimension, which corresponds to the negative ELBO or IWAE with log 2 base and normalized by data dimension, lower is better. In Figure a, we see that IWAE improves the ELBO when using classic amortized inference but is approximately equal to the ELBO when using optimal inference, which indicates the bound is tight. In Figure b, we compare the IWAE with classic amortized inference, optimal inference and the  the proposed reverse half-asleep (RHS) inference. Here we find the proposed method can also improve the classic IWAE estimation without training on the test data. In Figure 3, we plot the ELBO and IWAE for all three amortized inference methods.\label{fig:iwae}}
\end{figure*}

In this section we want to verify the tightness of the ELBO as a lower bound of the log likelihood. Consider the likelihood for a single data point $x'$, we have
\begin{align}
    \log p_\theta(x')\geq \langle\log p_\theta(x'|z)\rangle_{q_\phi(z|x')}-\mathrm{KL}(q_\phi(z|x')||p(z))\equiv \mathrm{ELBO}(x,\theta,\phi).
\end{align}
To evaluate $\log p_\theta(x')$, we can  use an importance weighted estimation (IWAE \cite{burda2015importance}), which can be rewritten as
\begin{align}
    \log p_\theta(x')=\log  \Big\langle \frac{p_\theta(x'|z)p(z)}{q_\phi(z|x)}\Big\rangle_{q_\phi(z|x)}\approx \log \frac{1}{K}\sum_{k=1}^K \frac{p_\theta(x'|z_k)p(z_k)}{q_\phi(z_k|x')}\equiv \mathrm{IWAE}_k(x,\theta,\phi),
\end{align}
where $z_k\sim q_\phi(z|x')$. The accuracy of the importance sampling heavily depends on the proposal distribution $q_\phi(z|x')$ and will be poor if $q_\phi(z|x')$ underestimates the high density region of $p_\theta(z|x)$~\cite{burda2015importance}. For the ELBO with optimal inference, we can assume the approximate posterior is close to the true posterior, so if the lower bound is tight,  we will observe that the ELBO is approximately equal to the IWAE. In Figure \ref{fig:iwae} we compare the ELBO and IWAE using classic amortized inference and optimal inference respectively (we use $k=10$ in all cases). We find that the IWAE can improve the ELBO for the traditional amortized inference and is approximately equivalent to the ELBO using the optimal inference strategy. Therefore, we can conclude that the ELBO with the optimal inference strategy is tight to $\log p_\theta(x)$.


We also estimate the IWAE using the proposal posterior learned by the proposed reverse half-asleep inference and find that our method can also improve the IWAE result, see Figure \ref{fig:iwae} for details. This is intuitive since our method can provide a better proposal distribution for importance sampling.

\section{Amortized Posterior for Down-stream Classification Task\label{app:representation}}
\begin{figure*}[h]
     \centering
      \begin{subfigure}[b]{0.45\textwidth}
         \centering
         \includegraphics[width=\textwidth]{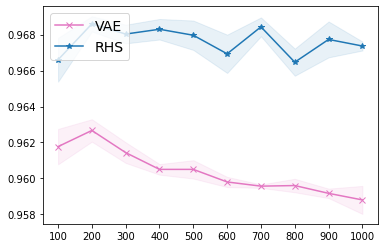}
         \caption{Binary MNIST}
     \end{subfigure}
     \begin{subfigure}[b]{0.45\textwidth}
         \centering
         \includegraphics[width=\textwidth]{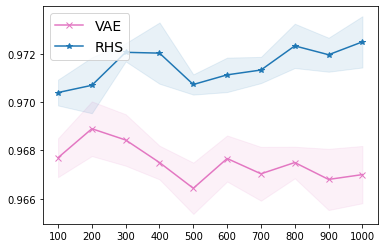}
         \caption{Grey MNIST}
     \end{subfigure}
        \caption{Representation Learning for Down-Stream Classification.\label{fig:rep} We train the VAE for 1000 epochs and  evaluate the classification accuracy (y-axis, higher is better) on the down-stream classification task every 100 epochs (x-axis). The results are averaged over 3 random seeds and we also plot the standard deviation.}
\end{figure*}

In Section 4, we discussed that the proposed reverse half-asleep method can improve the posterior prediction for the test data. One direct application is to use the learned amortized posterior $q_\phi(z|x)$ for down-stream tasks, e.g. image classification, where the samples $z'\sim q_\phi(z|x')$ can be treated as the `stochastic representation'~\cite{zhang2022improving,bengio2013representation} of the given data point $x'$. Given a labeled dataset $\{(x_1,y_1),\cdots, (x_N,y_N)\}$ and a trained amortized posterior (encoder) $q_\phi(z|x)$, we can then train a classifier $p_\eta(y|z)$ that maps from the latent space $z$ to the label $y$. 
After training the classifier, for a given test set of unlabelled data $\{x'_1,\cdots, x'_M\}$, the predictive distribution can be written as $p(y|x)=\int p_\eta(y|z)q_\phi(z|x)dz$ and can be approximated by Monte-Carlo: $p(y|x)\approx \frac{1}{K}\sum_{k=1}^K p(y|z'_k)$, where $z'_k\sim q_\phi(z|x)$. We train a classifier with 2 layer feed-forward neural network with hidden size 200, RelU activation and dropout with rate 0.1 on two datasets: binary MNIST and grey MNIST. The models are trained for 10 epochs with Adam optimizer and learning rate $3{\times}10^{-4}$. During training, we randomly sample one $z'$ for each data point $x$ and we use $k=100$ in the testing stage to estimate the predictive distribution. Figure \ref{fig:rep} shows the comparisons between the posterior trained by the classic amortized inference and the proposed reverse half-asleep method respectively. We can see our method consistently improves the classification accuracy performance.

\section{Effects of the Latent Space Dimensionality \label{app:dimensionality}}
\begin{figure*}[h]
     \centering
     \begin{subfigure}[b]{0.32\textwidth}
         \centering
        \includegraphics[width=\textwidth]{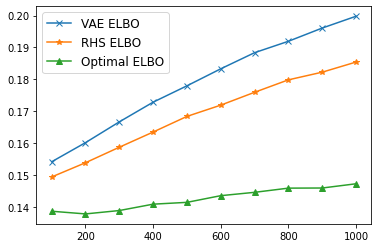}
         \caption{$\text{Dim}(z)=16$}
     \end{subfigure}
     \begin{subfigure}[b]{0.32\textwidth}
         \centering
        \includegraphics[width=\textwidth]{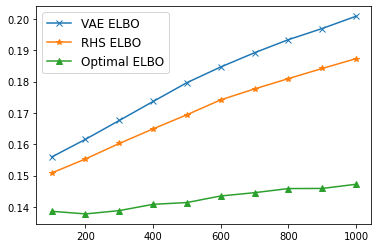}
         \caption{$\text{Dim}(z)=64$}
     \end{subfigure}
\begin{subfigure}[b]{0.32\textwidth}
         \centering
         \includegraphics[width=\textwidth]{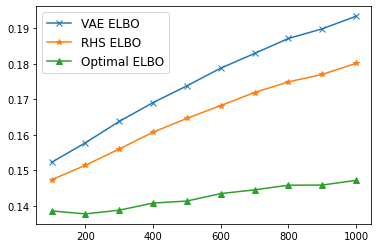}
         \caption{$\text{Dim}(z)=128$}
     \end{subfigure}
     \caption{Effects of different latent dimension. The y-axis is the BPD and x-axis is the training epochs. We find the amortized inference generalization gap exits in all  cases.\label{fig:latent:dimension}}
\end{figure*}

We study the effect of the latent dimension size on the generalization of the amortized inference. We use the VAE described in Section 4 with different latent size $[16,64,128]$ on Binary MNIST, see Figure \ref{fig:latent:dimension} for the result. We find the overfitting of amortized inference happens in all cases regardless of the latent size. We also apply the proposed reverse half-asleep training method to the saved model every 100 epoch and found our method can consistently improve the generalization performance.

\section{Reverse Half-asleep From the Beginning}
\begin{figure*}[h]
     \centering
     \begin{subfigure}[b]{0.45\textwidth}
         \centering
        \includegraphics[width=\textwidth]{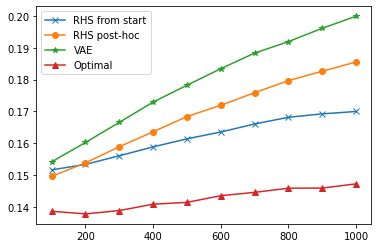}
         \caption{Binary MNIST}
     \end{subfigure}
     \begin{subfigure}[b]{0.45\textwidth}
         \centering
        \includegraphics[width=\textwidth]{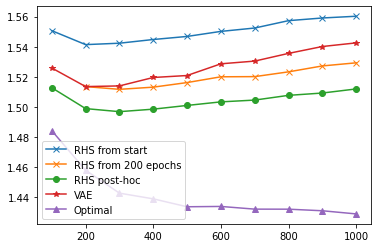}
         \caption{Grey MNIST}
     \end{subfigure}
     \caption{We compare different ways of using the proposed training objective (from the beginning or post hoc). We also plot the standard VAE training and the ELBO with optimal inference for reference.  \label{fig:begining}}
\end{figure*}

In section \ref{sec:experiments} we applied the reverse half-asleep training in a post-hoc fashion, which allow us to isolate the degree to which both
the generative model and amortized inference generalization gap are contributing to overfitting. It has also been observed that a poor variational posterior in the early stage of the training will cause the M-step of the generative model $p_\theta(x|z)$ to get trapped into a local minimum ( see ``Two problems with variational expectation maximization for
time-series models'' section in \cite{barber2011bayesian}). Therefore, we can also apply the proposed method from the beginning of training, see Figure~\ref{fig:begining} for the results. For a simple dataset like Binary MNIST, we find that using the proposed reverse half-asleep from the beginning can lead to a better test ELBO compared to the classic VAE training, or our proposed post-hoc training. However, for a more complex dataset like grey-scale MNIST, the result of using the reverse half-asleep from the beginning is worse than the classic VAE training. We hypothesize that for a complex dataset, the decoder in the beginning cannot generate valid images, which will lead to biased gradients. Therefore, we also report the result of using the reverse half-asleep training starting from 200 epochs onwards and find it is  better than the classic VAE training but is worse than the post-hoc reverse half-asleep method. We leave the study of how to improve the generalization from the beginning of the training to future work.

\section{Introduction of Bits Back Coding with VAE\label{app:bbans}}
In this section, we briefly introduce the background of the model-based lossless compression and the Bits Back coding scheme of VAE models. 
\subsection{Model-based Lossless Compression}
The goal of the lossless compression is to create an invertible mapping from real-world data (e.g. image, audio, video) to binary strings with the lengths of the strings as short as possible.

Let $X$ be a discrete random variable that taking values from a finite countable set $\gX$ and has a probability mass function (PMF) $p:\gX\rightarrow \sR$ such that $\forall x\in \gX$, $p(x)>0$ and $\sum_x p(x)=1$.
\begin{definition}
    The \textit{Shannon information content} of a sample $x\sim p(x)$ is defined as 
    \begin{align}
        h_p(x)\equiv-\log_2p(x).
    \end{align}
\end{definition}
\begin{definition}
    The \textit{Shannon Entropy} of a distribution $p$ is defined as 
    \begin{align}
        H(p)\equiv-\sum_xp(x)\log_2p(x).
    \end{align}
\end{definition}
We then give the informal statement of the \emph{Shannon Source Coding Theorem}~\cite{shannon1948mathematical}, the detailed statement and the proof can be found in Chapter 4 of \cite{mackay2003information}.
\begin{theorem}[Shannon's Source Coding Theorem (informal)]
     $N$ $i.i.d$ samples form the data generation distribution with PMF $p_d(x)$ can be losslessly compressed into more than $NH(p_d)$ bits when $N\rightarrow\infty$. Conversely, they cannot be losslessly compressed into fewer than $NH(p_d)$ bits.
\end{theorem}
To obtain a `near-optimal' lossless compression scheme in practice, one strategy is to compress each data $x\sim p_d(x)$ into a binary string with length equal to $h_{p_d}(x)+\epsilon$, where $h(x)$ is the \emph{Shannon information content} and $\epsilon$ represents a small coding overhead. Therefore, given $N$ $i.i.d$ samples $\{x_1,\cdots,x_N\}\sim p_d(x)$, the averaged compression length is
\begin{align}
    -\frac{1}{N}\sum_{n=1}^N \log_2 p_d(x_n)+\epsilon\xrightarrow{N\rightarrow+\infty} -\sum_xp_d(x)\log_2p_d(x)+\epsilon=H(p_d)+\epsilon,
\end{align}
which is close to optimal when $\epsilon$ is small. 

Different coders are proposed to make the overhead $\epsilon$ for different types of data. For multi-dimensional data, there exits
two methods that can provide us `near-optimal' lossless compression: Arithmetic Coding (AC)~\cite{witten1987arithmetic} and Asymmetric Numeral System (ANS)~\cite{duda2013asymmetric}, we recommend  Chapter 6 of~\cite{mackay2003information} and \cite{townsend2020tutorial} for detailed introductions of the two methods respectively. 
We use the ANS coder in this paper since it has a faster speed comparing to AC. For simplicity, we abstract an ANS coder as an invertible function $\mathrm{enc}_p(\cdot)$ that maps a given data $x'\in \mathcal{X}$ to a binary string message $m'$ with length $\mathrm{len}(m')=-\log_2 p(x')+\epsilon$, where $\epsilon$ is a negligible coding overhead. We also denote the decoding function as $\mathrm{dec}_p(\cdot)=\mathrm{enc}_p^{-1}(\cdot)$ and have $\mathrm{dec}_p(m')\rightarrow x'$.  

We have introduced how to optimally compress the data when we know the true data generation distribution $p_d(x)$. However, the distribution $p_d(x)$ is usually unknown in practice, we would like to learn a model $p_\theta(x)$ to approximate the underlying data distribution $p(x)$ and then use the learned model $p_\theta(x)$ to conduct lossless compression. In this case, the \emph{averaged} data compression length for $\{x_1,\cdots,x_N\}\sim p_d(x)$ is (ignoring the coding overhead $\epsilon$):
\begin{align}
    -\frac{1}{N}\sum_{n=1}^N\log_2 p_\theta(x_n)\xrightarrow{N\rightarrow+\infty} -\sum_x p(x)\log_2 p_\theta(x_n).
\end{align}
The difference  between the model compression length and the optimal compression length is 
\begin{align}
    -\frac{1}{N}\sum_{n=1}^N \Big(\log_2 p_\theta(x_n)-\log_2 p_d(x_n)\Big)\xrightarrow{N\rightarrow+\infty} \KL(p_d(x)||p_\theta(x)).
\end{align}

\subsection{Bits Back Compression with VAEs}
Given a discrete Latent VAE model specified by the PMFs $\{p_\theta(x|z),q_\phi(z|x),p(z)\}$ and a target data $x'$ to compress. A naive strategy is to first generate a sample $z'\sim q_\phi(z|x')$ and then encode $x'$ with $p_\theta(x|z')$. We also encode $z'$ with distribution $\log p(z)$, so the total code length is then  We also encode $z'$ with distribution $\log p(z)$, so the total code length is then 
\begin{align}
    -\log_2 p_\theta(x'|z')-\log_2 p(z'),
\end{align}
which is larger than the optimal code length $-\log_2 p(x')$ by $-\log_2 p_\theta(z'|x')$ bits. To achieve the optimal code length, a key observation is that the sampling process $z'\sim q_\phi(z|x')$ can be done by decoding random bits using the distribution $q_\phi(z|x')$. 
Specifically, we assume that we can access a message that already contains random bits, which we visualize as the following figure\footnote{The visualization is taken from~\cite{townsend2019practical}.}.

\newlength{\messageheight}
\setlength{\messageheight}{8pt}
\begin{tikzpicture}
  \draw (3, \messageheight) -- (0, \messageheight) -- (0, 0) -- (3, 0);
  \draw(3, \messageheight) -- (3, 0);

   \node [label=above:\small{\text{Initial random bits}}] at (1.5, \messageheight) {};
\end{tikzpicture}

\textbf{In the encoding stage}, we first sample $z'$ form $q_\phi(z|x')$  by decoding random bits with distribution $q_\phi(z'|x')$, so the message length decreased by length $-\log_2q_\phi(z'|x')$.

\begin{tikzpicture}
  \draw (1.5, \messageheight) -- (0, \messageheight) -- (0, 0) -- (1.5, 0)--(1.5, \messageheight);
  \draw[dashed]   (1.5, \messageheight) -- (1.5, 0) -- (3, 0)
    -- (3, \messageheight) -- cycle;

  \draw (1.5, \messageheight + 10) -- (3, \messageheight + 10);
  \draw (1.5, \messageheight + 13) -- (1.5, \messageheight + 7 );
  \draw (3, \messageheight + 13) -- (3, \messageheight + 7 );
  \node [label=above:\small{\(-\log_2 q_\phi(z'|x')\)}] at (2.25, \messageheight + 10) {};
\end{tikzpicture}$\quad \xrightarrow{\mathrm{dec}_{q_\phi(z|x')}(\cdot)}\quad z'$

We then encode $x'$ with distribution $p_\theta(x'|z')$, so the message is increased by length $-\log_2 p_\theta(x'|z')$.
 \begin{tikzpicture}
  \draw (2.5, 0) -- (0, 0) -- (0, \messageheight) -- (2.5, \messageheight);
  \draw (1.2, 0) -- (1.2, \messageheight);
  \draw  (2.5, 0) -- (2.5, \messageheight);
  \draw (1.2, \messageheight + 10) -- (2.5, \messageheight + 10);
  \draw (1.2, \messageheight + 13) -- (1.2, \messageheight + 7 );
  \draw (2.5, \messageheight + 13) -- (2.5, \messageheight + 7 );
  \node [label=above:\small{\(-\log_2p_\theta(x'|z')\)}]
    at (1.85, \messageheight + 10) {};
  \end{tikzpicture}$\quad \xleftarrow{\mathrm{enc}_{p_\theta(x|z')}(\cdot)}\quad x'$

Finally, we encode  $z'$ with distribution $p(z)$ and the message length is increased by $-\log p(z')$.

      \begin{tikzpicture}
  \draw (4.1, 0) -- (0, 0) -- (0, \messageheight) -- (4.1, \messageheight);
  \draw (1.2, 0) -- (1.2, \messageheight);
  \draw (2.5, 0) -- (2.5, \messageheight);
  \draw  (4.1, 0) -- (4.1, \messageheight);
  \draw (2.5, \messageheight + 10) -- (4.1, \messageheight + 10);
  \draw (2.5, \messageheight + 13) -- (2.5, \messageheight + 7 );
  \draw (4.1, \messageheight + 13) -- (4.1, \messageheight + 7 );
  \node [label=above:\small{\(-\log_2 p(z')\)}]
    at (3.3, \messageheight + 10) {};
  \end{tikzpicture}$\quad \xleftarrow{\mathrm{enc}_{p(z)}(\cdot)}\quad z'$

\textbf{In the decoding stage}, we first decode $z'$ using $p(z)$.

      \begin{tikzpicture}
  \draw (2.5, 0) -- (0, 0) -- (0, \messageheight) -- (2.5, \messageheight) -- (2.5, 0) ;
   \draw[dashed] (2.5, 0) -- (4.1, 0) -- (4.1, \messageheight) -- (2.5, \messageheight)-- cycle;
  \draw (1.2, 0) -- (1.2, \messageheight);
  \draw    (2.5, \messageheight + 10) -- (4.1, \messageheight + 10);
  \draw (2.5, \messageheight + 13) -- (2.5, \messageheight + 7 );
  \draw (4.1, \messageheight + 13) -- (4.1, \messageheight + 7 );
  \node [label=above:\small{\(-\log_2 p(z')\)}]
    at (3.3, \messageheight + 10) {};
  \end{tikzpicture}$\quad \xrightarrow{\mathrm{dec}_{p(z)}(\cdot)}\quad z'$

We then decode $x'$ with distribution $p_\theta(x|z')$.

   \begin{tikzpicture}
  \draw (1.2, 0) -- (0, 0) -- (0, \messageheight) -- (1.2, \messageheight)-- (1.2, 0) ;
  \draw[dashed] (1.2, 0) -- (2.5, 0) -- (2.5, \messageheight) -- (1.2, \messageheight);

  \draw (1.2, \messageheight + 10) -- (2.5, \messageheight + 10);
  \draw (1.2, \messageheight + 13) -- (1.2, \messageheight + 7 );
  \draw (2.5, \messageheight + 13) -- (2.5, \messageheight + 7 );
  \node [label=above:\small{\(-\log_2p_\theta(x'|z')\)}]
    at (1.85, \messageheight + 10) {};
  \end{tikzpicture}$\quad \xrightarrow{\mathrm{dec}_{p_\theta(x|z')}(\cdot)}\quad x'$

Finally, we encode the random bits `back' to the stack to recover the initial message.

\begin{tikzpicture}
  \draw (1.5, \messageheight) -- (0, \messageheight) -- (0, 0) -- (1.5, 0)--(1.5, \messageheight);
  \draw  (1.5, \messageheight) -- (1.5, 0) -- (3, 0)
    -- (3, \messageheight) -- cycle;

  \draw (1.5, \messageheight + 10) -- (3, \messageheight + 10);
  \draw (1.5, \messageheight + 13) -- (1.5, \messageheight + 7 );
  \draw (3, \messageheight + 13) -- (3, \messageheight + 7 );
  \node [label=above:\small{\(-\log_2 q_\phi(z'|x')\)}] at (2.25, \messageheight + 10) {};
\end{tikzpicture}$\quad \xleftarrow{\mathrm{enc}_{q_\phi(z|x')}(\cdot)}\quad z'$

Therefore, the `net' message length to compress data $x'$ with a VAE is equivalent to 
\begin{align}
    -\log_2 p_\theta(x|z')-\log_2p(z')+\log q_\phi(z'|x'),
\end{align}
which is a one-sample estimation of the ELBO and is optimal (equal to $-\log_2 p_\theta(x')$) when the amortized variational posterior is equal to the true posterior $q_\phi(z|x')=p_\theta(z|x')$. 

This scheme and also be extended to continuous latent $z$ with negligible cost by 
quantizing the PDF $p(z)$ and $q_\phi(z)$ into PMF to conduct the compression. See \cite{townsend2019practical}  for details.
This `Bits Back' coding method was first introduced as a thought experiment in~\cite{wallace1990classification,hinton1993keeping} and was later implemented by~\cite{frey1996free} with an AC coder.  Recently, \cite{townsend2019practical} proposed to implement the Bits Back with ANS~\cite{townsend2019hilloc} and a VAE model, which allows great improvement of both the compression rate and the computational efficiency. We refer the reader to~\cite{townsend2019practical} for other practical considerations and implementation details.

\end{document}

%% file: math_commands.tex
\usepackage{amsmath,amsfonts,bm}

\newtheorem{theorem}{Theorem}

\newtheorem{definition}{Definition}
\newcommand*\ruleline[1]{\par\noindent\raisebox{0.1cm}{\makebox[\linewidth]{\hrulefill\hspace{0.8ex}\raisebox{-.5ex}{#1}\hspace{0.8ex}\hrulefill}}}

\def\eqref#1{equation~\ref{#1}}

\def\1{\bm{1}}

\DeclareMathAlphabet{\mathsfit}{\encodingdefault}{\sfdefault}{m}{sl}
\SetMathAlphabet{\mathsfit}{bold}{\encodingdefault}{\sfdefault}{bx}{n}

\def\gX{{\mathcal{X}}}

\def\sR{{\mathbb{R}}}

\newcommand{\KL}{{\mathrm{KL}}}

